\documentclass[10pt,twocolumn,letterpaper]{article}
\usepackage[T1]{fontenc}
\usepackage{paper}
\usepackage{graphicx}
\usepackage{amsmath,amssymb}
\usepackage{booktabs}
\usepackage{tabularx}
\usepackage{enumitem}
\usepackage{dblfloatfix}
\usepackage{flafter}
\usepackage{placeins}
\usepackage{float}
\usepackage{microtype}
\usepackage{xcolor}
\usepackage{url}
\definecolor{linkblue}{rgb}{0.13,0.35,0.60}
\usepackage[breaklinks,colorlinks,allcolors=linkblue]{hyperref}
\hypersetup{
  pdftitle={Tiny Dreamer: Lightweight World Models for RGB UAV Control},
  pdfauthor={Mengzhao Li, Zijing Zhang, Ruixin Wang, Muhan Lyu, Jifeng Hu, Jinwu Gao, Huajun Chen, Yi Chang, Hechang Chen},
  pdfsubject={arXiv preprint},
  pdfkeywords={world models, UAV control, actuator-conditioned imagination, dynamic routing}
}
\newcommand{\papertablesetup}{\small\setlength{\tabcolsep}{6pt}\renewcommand{\arraystretch}{1.16}}
\newsavebox{\papertablebox}
\newenvironment{checkedtable}{\begin{lrbox}{\papertablebox}}{%
  \end{lrbox}%
  \ifdim\wd\papertablebox>\dimexpr\linewidth+0.5pt\relax
    \PackageError{TinyDreamer}{Table exceeds the available column width}{Break long headings or adjust column widths; do not shrink the document font.}%
  \fi
  \usebox{\papertablebox}}
\title{Tiny Dreamer: Lightweight World Models for RGB UAV Control}
\author{Mengzhao Li\textsuperscript{*}, Zijing Zhang\textsuperscript{*}, Ruixin Wang\textsuperscript{*}, Muhan Lyu\textsuperscript{*}, Jifeng Hu\textsuperscript{*}, Jinwu Gao\textsuperscript{*}, Huajun Chen\textsuperscript{$\circ$},\\
Yi Chang\textsuperscript{*}, Hechang Chen\textsuperscript{*,2}\\[8pt]
{\normalsize \textsuperscript{*}Jilin University \quad
 \textsuperscript{$\circ$}Zhejiang University}}
\date{}
\AtBeginDocument{%
  \setlength{\abovedisplayskip}{9pt plus 2pt minus 2pt}%
  \setlength{\belowdisplayskip}{9pt plus 2pt minus 2pt}%
  \setlength{\abovedisplayshortskip}{4pt plus 1pt minus 1pt}%
  \setlength{\belowdisplayshortskip}{5pt plus 1pt minus 1pt}%
  \setlength{\jot}{4pt}%
}
\begin{document}
\maketitle
\flushbottom
\setlength{\parskip}{0pt plus 4pt minus 0pt}

\begin{abstract}
Vision-based UAV control requires temporal reasoning within limited onboard computing budgets, while actuator mismatch complicates the prediction of action outcomes.
We present Tiny Dreamer, a lightweight world-model framework designed to reduce online computation and improve policy learning under actuator variation.
A Dynamic Routing Network (DRNet) adaptively activates directional, local, and spectral branches with low-rank channel mixing, while a causal Transformer summarizes RGB and command history.
Actuator-Conditioned Imagination (ACI) conditions a few-step latent generator on thrust gain, response lag, and command delay, keeping the actuator domain fixed throughout each imagined rollout.
We use imagined rollouts from the world model to train the actor and critic.
During flight, the visual encoder and causal Transformer process RGB and command history, and the actor issues commands to the low-level flight controller, which stabilizes the UAV.
Experiments show that routing reduces active computation and inference latency while retaining competitive control performance.
Actuator conditioning improves long-horizon prediction and reduces tracking error under the evaluated shifts, with zero-shot trajectory tracking demonstrated on a physical UAV.
Dynamic routing reduces active computation by 32.1\% and P95 inference latency by 13.9\% relative to all-branch execution, and scheduled ACI reduces zero-shot tracking RMSE on the P450 by 17.0\% relative to unconditioned actuator randomization.
\end{abstract}

\begin{figure}[!t]
\centering
\includegraphics[width=\columnwidth]{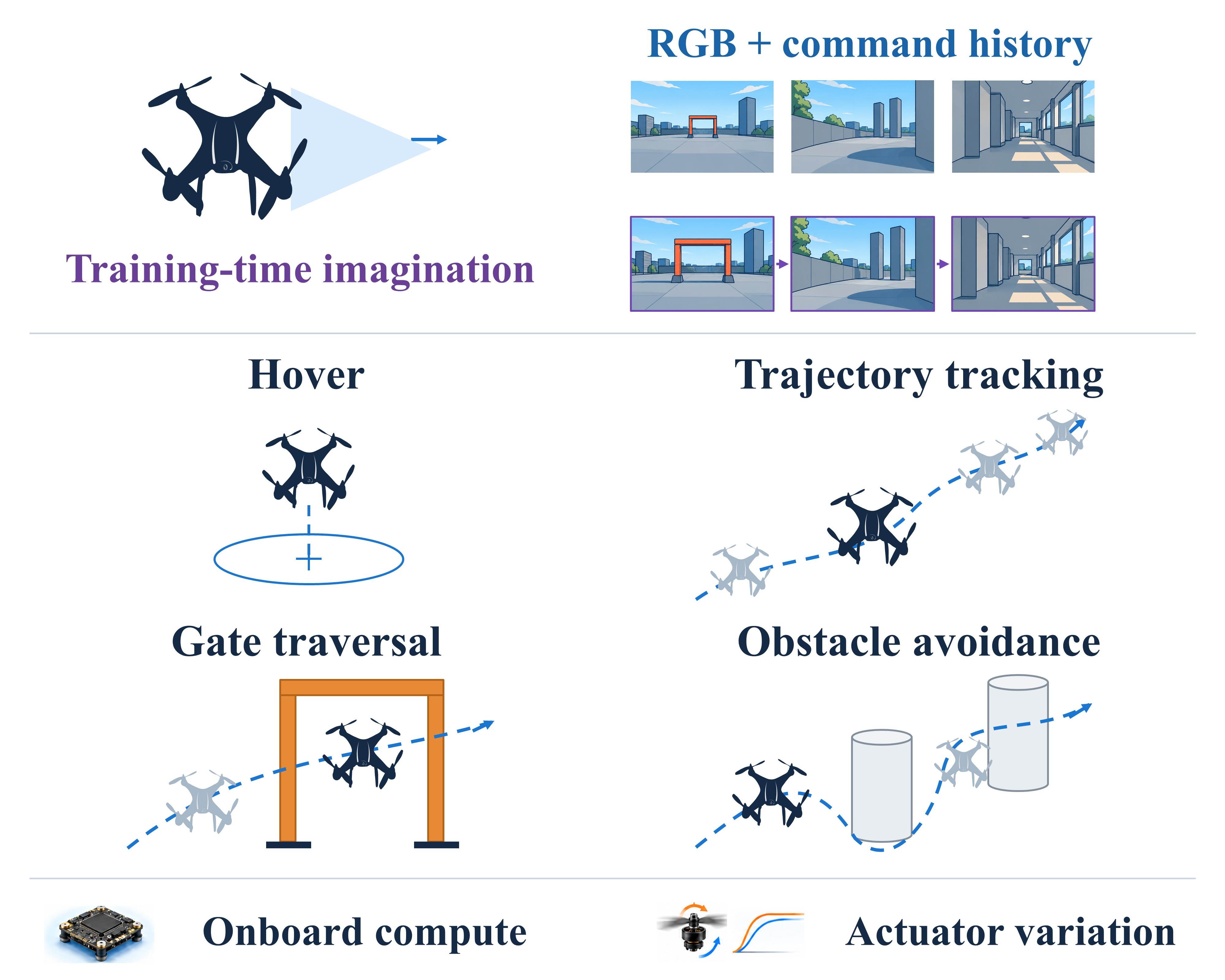}
\caption{\textbf{Tiny Dreamer for visual UAV control.}
RGB and command history support four control tasks under onboard computing constraints and actuator variation.}
\label{fig:intro_overview}
\end{figure}

\section{Introduction}
\footnotetext[2]{Correspondence: chenhc@jlu.edu.cn, School of Artificial Intelligence, Jilin University.}
Vision-based unmanned aerial vehicle (UAV) control must infer motion from images and issue timely commands within onboard computing budgets.
Blur, occlusion, and actuator-response ambiguity make temporal reasoning essential, while inference delays can render observations stale.
Prior learned-flight systems~\cite{swift,dreamtofly,skydreamer} motivate efficient models with explicit sensing and command interfaces.
Figure~\ref{fig:intro_overview} summarizes our tasks and deployment challenges.
In Tiny Dreamer, the RGB-only policy receives RGB frames as its sole sensory input, along with its own past control commands.
The low-level flight controller uses state feedback for stabilization.

Model-based reinforcement learning uses a learned world model to predict future outcomes conditioned on actions and trains the policy on imagined rollouts~\cite{worldmodels,planet,dreamer,dreamerv3}.
Prediction errors can accumulate along these rollouts and bias policy learning, particularly when the world model fails to capture changes in actuator response.
The resulting problem is to learn from actuator-consistent imagined futures while keeping deployment within onboard inference budgets.
Tiny Dreamer conditions imagined rollouts on actuator parameters so that the policy can train across different thrust gains, response lags, and command delays.
To reduce onboard computation, Tiny Dreamer performs
imagined rollouts only during training, retaining
the visual encoder, finite-history model, and actor
for deployment.

Executing every visual branch imposes fixed computation,
even when different observations require different
visual features.
Although efficient backbones~\cite{mobilenetv2,mobilenetv3,efficientnet}
reduce overall cost, their execution paths remain fixed.
Conditional computation~\cite{skipnet,condconv,dynamicconv}
enables adaptive execution, but efficiently selecting
complementary spatial and spectral features remains
challenging.
To address this, we propose DRNet, which adaptively
selects directional, local, and spectral branches with
low-rank channel mixing.
The spectral branch applies learnable frequency-band
filtering through FFT and inverse FFT.
Because routing and frequency transforms introduce
overhead, we evaluate both active computation
and measured device latency.

Differences in thrust gain, response lag, and command delay can cause the same control command to produce different UAV motions.
Dynamics randomization samples different combinations of these parameters during training, exposing the policy to varied actuator responses~\cite{dynamicsrandomization}.
When training a policy on imagined rollouts, the world model must also predict action outcomes under these different parameter combinations.
ACI therefore conditions the latent generator on actuator parameters and keeps them fixed throughout each imagined rollout, so that the generated trajectory reflects a consistent actuator response.
During flight, the actor receives no actuator parameters and acts from RGB and command history.

Tiny Dreamer combines DRNet and ACI in a controller with a fixed history window.
We train the controller in Isaac Sim and evaluate its zero-shot transfer to Gazebo and a physical P450 quadrotor.
Dynamic routing reduces computation and inference latency while achieving a gate traversal success rate comparable to executing all visual branches.
In real-world trajectory tracking, scheduled ACI achieves lower tracking error than unconditioned actuator randomization.

Our key contributions are as follows:
\begin{itemize}[leftmargin=1.2em,itemsep=0pt,topsep=2pt]
\item We propose DRNet, a structured visual encoder with conditional directional, local, and spectral branches and low-rank channel mixing, designed to skip inactive branches during inference.
\item We formulate ACI, which conditions latent future generation on thrust gain, response lag, and command delay while holding the actuator condition consistent across each imagined trajectory.
\item We integrate DRNet and ACI into a finite-context UAV controller and evaluate task performance, onboard inference cost, and zero-shot transfer in Gazebo and on a physical P450 quadrotor.\end{itemize}

\begin{figure*}[!t]
\centering
\includegraphics[width=0.95\textwidth]{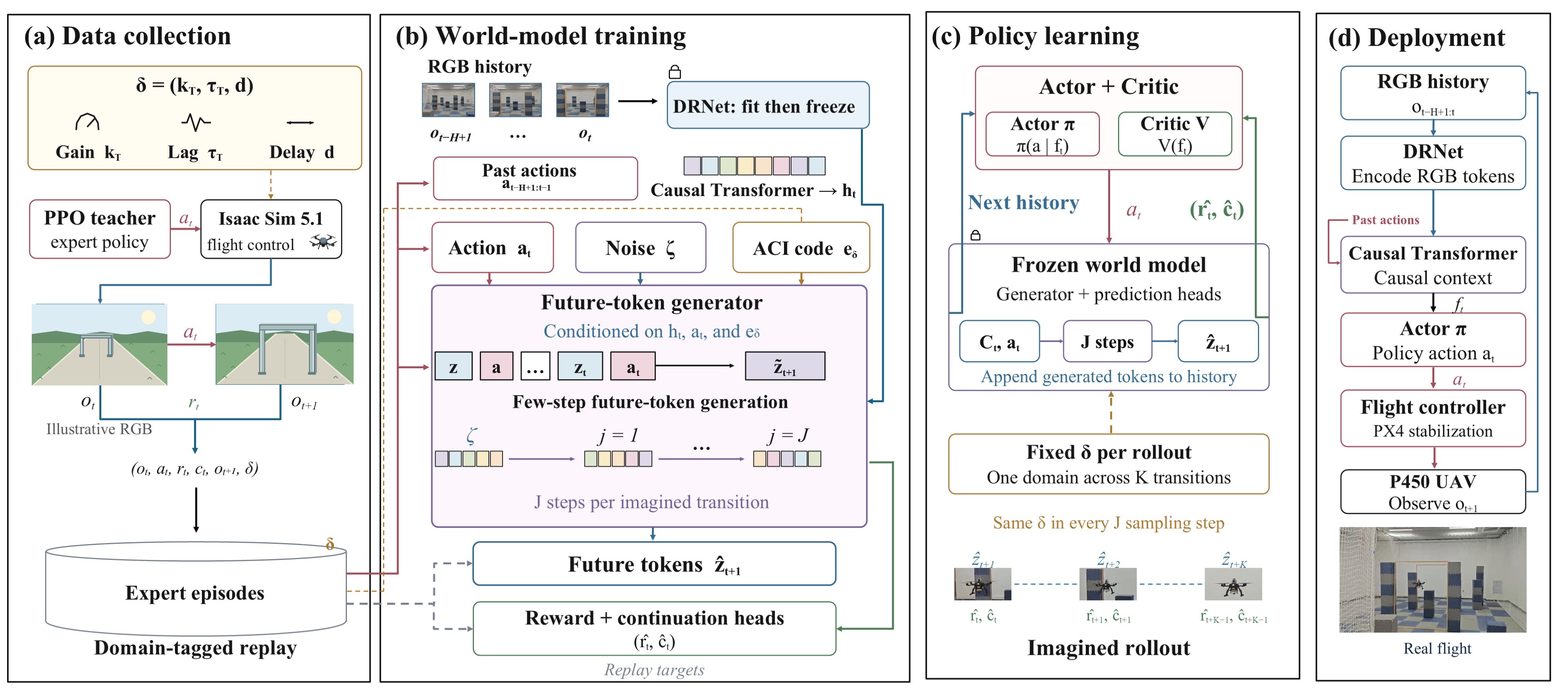}
\caption{\textbf{Tiny Dreamer training and deployment.} (a) Domain-tagged data collection with a PPO teacher~\cite{ppo}. (b) Routed encoding and actuator-conditioned latent prediction. (c) Actor--Critic learning with a fixed actuator domain per imagined rollout. (d) Deployment: DRNet encodes RGB observations, the causal Transformer processes visual tokens and action history, and the actor outputs commands to the flight controller.}
\label{fig:framework}
\end{figure*}

\section{Related Work}
Visual UAV control connects three questions: how to predict useful futures from images, how to keep visual inference within an onboard budget, and how to learn when command-to-motion dynamics vary. We organize prior work around these questions and use them to clarify the scope of Tiny Dreamer.

\textbf{World Models and Visual Flight.} World Models~\cite{worldmodels}, PlaNet~\cite{planet}, and the Dreamer family~\cite{dreamer,dreamerv2,dreamerv3,dreamer4} established latent prediction and imagined behavior learning for visual control; DreamerV2 extended this line with discrete latent states~\cite{dreamerv2}. Subsequent approaches address different uses of the learned model: PETS represents prediction uncertainty, TD-MPC and TD-MPC2 optimize control trajectories, and EfficientZero combines model learning with search~\cite{pets,tdmpc,tdmpc2,efficientzero}. IRIS, STORM, and DIAMOND develop temporal or generative world models~\cite{iris,storm,diamond}, while flow matching, rectified flow, consistency models, and shortcut models motivate efficient latent generation~\cite{flowmatching,rectifiedflow,consistency,shortcut}. In visual flight, Swift, Dream to Fly~\cite{dreamtofly}, SkyDreamer~\cite{skydreamer}, and AirDreamer~\cite{airdreamer} study learned UAV policies with differing sensing and execution interfaces; AirDreamer also uses a world model~\cite{airdreamer}. These lines motivate our use of imagined futures; our focus is to keep each rollout consistent with one actuator response while measuring the deployed policy's visual cost.

\textbf{Visual Computation Under Onboard Budgets.} MobileNetV2/V3~\cite{mobilenetv2,mobilenetv3} and EfficientNet~\cite{efficientnet} reduce the cost of a fixed visual backbone. Conditional computation instead varies the work performed for each input: Squeeze-and-Excitation recalibrates channels~\cite{senet}, CondConv and Dynamic Convolution mix kernels, and SkipNet and sparse experts adapt execution paths~\cite{moe}. DrQ improves learning from pixels through augmentation~\cite{drq}, addressing data use rather than the cost of an executed policy. DRNet selects among directional, local, and spectral operators and applies low-rank channel mixing. Because selection, Fourier transforms, and memory traffic carry overhead, we report active GFLOPs alongside measured device latency.

\textbf{Actuator Variation and Sim-to-Real Transfer.} Domain randomization exposes policies to changes in appearance and physical response; SimOpt adapts simulation distributions using real experience~\cite{domainrandomization,simopt}. Meta-learned dynamics, CaDM, and PEARL infer context from experience~\cite{metaadapt,cadm,pearl}, while FiLM conditions features~\cite{film}. These methods motivate adaptation to uncertain dynamics. ACI uses known thrust gain, response lag, and command delay to condition training-time future generation, holding one condition fixed throughout each imagined rollout. At deployment, the generator is removed and the actor receives neither actuator codes nor parameter updates; it acts from RGB and command history.

\section{Method}
\label{sec:method}
Figure~\ref{fig:framework} shows training; $\theta,\phi,\omega,\eta,\psi,\xi$ denote the encoder, history model, generator, actuator adapter, actor, and critic.

\subsection{Observation and Action Interface}
\label{sec:interface}
Visual flight is a partially observed control problem maximizing $\mathbb E_\pi\sum_t\gamma^t r_t$.
The actor observes $64\times64$ RGB images and its issued commands:
\begin{equation}
\begin{split}
z_t&=f_\theta(o_t)\\
h_t&=g_\phi(z_{t-H+1:t},a_{t-H+1:t-1})\\
a_t&\sim\pi_\psi(\cdot\mid h_t).
\end{split}
\label{eq:actor_chain}
\end{equation}
Here $f_\theta$, $g_\phi$, and $\pi_\psi$ are the visual encoder, history model, and actor.
Context has $H$ images and $H-1$ actions with masked reset padding; bounded body-frame velocity/yaw-rate commands are clipped before replay.
Depth, gate labels, actuator parameters, and external state estimates are excluded from the actor.
FAST-LIO and PX4 provide low-level feedback and stabilization; simulator state or recorded trajectories provide rewards and evaluation targets.
Observation, command, and actuator ticks are timestamp-aligned with frame holds.

\subsection{Structured Encoding and Sparse Routing}
\label{sec:encoder}
\begin{figure}[!t]
\centering
\includegraphics[width=\columnwidth]{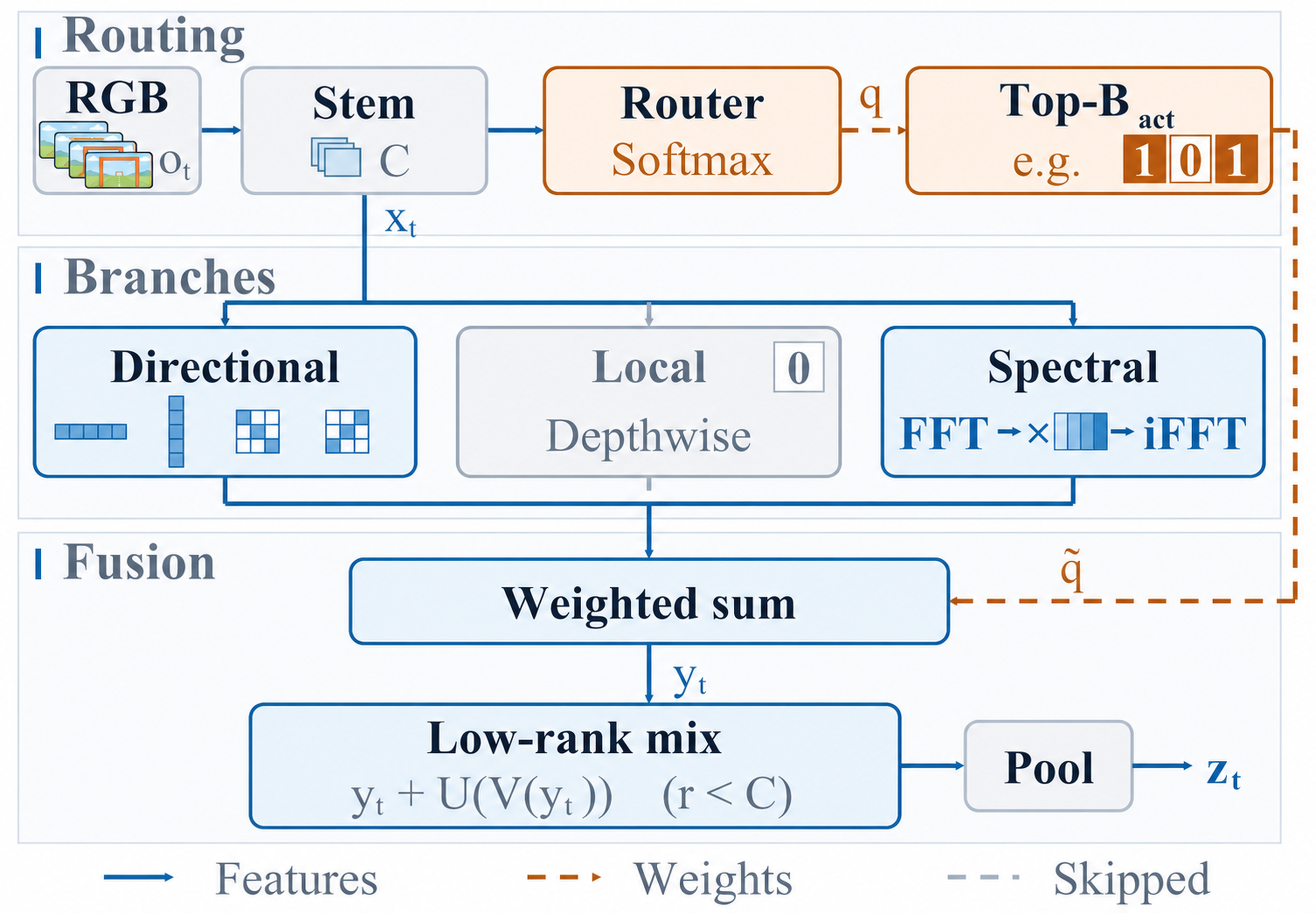}
\caption{\textbf{DRNet architecture.}
A shared stem feeds three operators; top-$B_{\mathrm{act}}$ routing reweights selected outputs before low-rank residual mixing and pooling.
Features flow along blue paths; dashed orange paths carry routing weights.}
\label{fig:drnet_structure}
\end{figure}

A shared stem produces $x_t$ with $C$ channels.
Figure~\ref{fig:drnet_structure} illustrates $B=3$ shape-compatible branches: directional, local depthwise, and spectral, motivated by efficient convolution and spectral mixing~\cite{mobilenet,shufflenet,fnet}.

\textbf{Learnable Directional Depthwise Convolution.} Four operators $D_i$ use horizontal $1\times5$, vertical $5\times1$, and main-/anti-diagonal masked $3\times3$ kernels, with $\mathrm{groups}=C$, unit stride, and size-preserving padding.
Their outputs are fused with learned weights:
\begin{equation}
\begin{split}
F_{\mathrm{dir}}(x_t)&=\sum_{i=1}^{4}\alpha_iD_i(x_t)\\
\alpha_i&=\operatorname{softmax}(\beta)_i.
\end{split}
\label{eq:dir_fusion}
\end{equation}
The masks are fixed, and the kernels and fusion logits $\beta$ are learned end-to-end.
Dense masked convolutions are costed as executed kernels, without assuming savings from zeroed weights.

\textbf{\mbox{Channel-Wise Frequency-Band Filtering.}} A learned spectral mask multiplies the RFFT2 coefficients of each channel:
\begin{equation}
M_c(u,v)=1+\sum_{b=1}^{B_f}\theta_{c,b}\phi_b(u,v).
\label{eq:freq_mask}
\end{equation}
The mask combines $B_f=4$ fixed real frequency-band bases, symmetric under frequency inversion, with learned per-channel gains $\theta_{c,b}$.
The real, inversion-symmetric response retains all frequencies and freezes at deployment; IRFFT2 restores the original spatial size, while routing remains dynamic.

The router scores each branch, a top-$B_{\mathrm{act}}$ mask $m_b$ keeps only its active entries, and the retained weights are renormalized:
\begin{equation}
\begin{split}
q_b&=\operatorname{softmax}(R(x_t))_b\\
\widetilde q_b&=\frac{m_bq_b}{\sum_jm_jq_j}\\
y_t&=\sum_b\widetilde q_bF_b(x_t).
\end{split}
\label{eq:routing}
\end{equation}
Active outputs are combined into $y_t$ and encoded by $z_t=\operatorname{Pool}(y_t+U(Vy_t))$.
Channel projections $V:\mathbb R^C\rightarrow\mathbb R^r$ and $U:\mathbb R^r\rightarrow\mathbb R^C$, with $r<C$, cost $2Cr$ per spatial location.
Initial soft routing transitions to a straight-through sparse mask.
Deployment must skip inactive branches to realize compute savings.

Codec training combines three terms:
\begin{equation}
\begin{split}
\mathcal L_{\mathrm{rec}}&=\mathbb E\|d_\kappa(z_t)-o_t\|_1\\
\mathcal L_{\mathrm{bal}}&=B\sum_b(\mathbb E[q_b]-1/B)^2\\
\mathcal L_{\mathrm{cost}}&=\mathbb E\sum_bq_bC_b.
\end{split}
\label{eq:codec_loss}
\end{equation}
The three terms target reconstruction, utilization balancing, and expected cost, and training then performs hard-route adaptation and freezing.
Here $C_b$ estimates branch cost, distinct from executed compute and device latency.
The decoder is training-only; reconstruction alone does not guarantee control sufficiency.

\subsection{Finite-Context Dynamics and Supervision}
\label{sec:dynamics}
Interleaved visual and past-action tokens are processed by a causal Transformer~\cite{transformer} to capture temporal dependencies between visual observations and the control commands issued by the policy.
Specifically, the Transformer operates on a finite context consisting of $H$ visual observations and $H-1$ past actions, preserving their temporal ordering.
The hidden representation associated with the latest visual token is used as the current history state $h_t$, which summarizes the available visual and action context for subsequent prediction.
The selected action $a_t$ is then provided to the future prediction module, allowing the latent dynamics to model the next-state transition conditioned on both the recent interaction history and the current control command.

To preserve temporal causality, the attention mask restricts each token to attending only to itself and preceding tokens, thereby preventing information leakage from future observations or actions.
With $2H-1$ interleaved tokens, standard full self-attention incurs quadratic computational complexity with respect to the context length.
Grouped-query attention (GQA)~\cite{gqa} and key-value caching may offer opportunities to reduce inference overhead; however, their actual benefits depend on the implementation and target hardware and therefore require separate profiling.
The fixed-context formulation also provides a consistent temporal interface for replay-based model training and online policy inference.

For latent dynamics supervision, the frozen visual codec provides a stop-gradient target representation of the next observation, where $\operatorname{sg}(\cdot)$ denotes the stop-gradient operator:
\begin{equation}
\bar z_{t+1}=\operatorname{sg}\bigl(f_\theta(o_{t+1})\bigr).
\label{eq:latent_target}
\end{equation}
This target supplies a visual representation for supervising future latent prediction without introducing external state estimates into the visual representation.
Notation separates the three quantities involved: the non-detached encoding $z_{t+1}=f_\theta(o_{t+1})$ of the real next observation, its stop-gradient copy $\bar z_{t+1}$, which serves as the regression target, and the generator prediction $\hat z_{t+1}$.
In parallel, the reward and continuation heads are supervised using replay transitions, with $(h_t,a_t,z_{t+1})$ as their inputs, so both heads are trained on real encoded latents rather than on generated ones.
The reward head estimates the immediate task reward, whereas the continuation head predicts whether the transition continues beyond the current step.
Time-limit truncation is distinguished from task termination so that an episode ending due to a predefined time horizon is not incorrectly treated as a terminal event.
Regarding task specification, hover and trajectory-tracking tasks use task-specific fixed references, whereas gate traversal and obstacle avoidance rely on visual targets.
These references are used in task definition and evaluation rather than supplied as arbitrary reference inputs to the policy.
Consequently, the deployed controller operates from RGB observations and its own command history without requiring an additional arbitrary-reference input.

\subsection{Few-Step Conditional Future Generation}
\label{sec:generator}
Motivated by flow matching~\cite{flowmatching}, shortcut models~\cite{shortcut}, and Dreamer~4~\cite{dreamer4}, we introduce a conditional clean-latent generator to predict future latent representations from the current observation history, previous actions, and actuator conditions.
Specifically, the conditioning context is defined as $c_t=(h_t,a_t,e_\delta)$, where $h_t$ denotes the encoded observation history, $a_t$ represents the corresponding action context, and $e_\delta$ encodes the actuator-domain parameters.
When actuator-conditioned imagination (ACI) is disabled, a fixed nominal code is used instead.

Given a Gaussian noise sample $\zeta\sim\mathcal N(0,I)$ and a target latent $\bar z_{t+1}$, we construct an interpolation between the noise distribution and the clean latent representation:
\begin{equation}
\begin{aligned}
x_s&=(1-s)\zeta+s\bar z_{t+1}\\
v_\omega(x_s,s,\ell;c_t)&=\frac{F_\omega(x_s,s,\ell;c_t)-x_s}{1-s}.
\end{aligned}
\label{eq:clean_velocity}
\end{equation}
Here $s\in[0,1)$ denotes the interpolation time, $\ell$ specifies the requested generation step, and $F_\omega$ predicts the corresponding clean latent endpoint.
The velocity parameterization therefore expresses the displacement from the current interpolated state to the predicted clean endpoint, normalized by the remaining interval $1-s$.

To support multi-step generation, $\ell$ is sampled from a predefined dyadic grid, while $s$ is selected from compatible grid points satisfying $s+\ell\leq1$.
For the minimum generation step $\ell_{\min}$, the observed clean latent $\bar z_{t+1}$ provides the direct training target.
For larger generation steps, we construct a shortcut target using two successive half-step evaluations of a periodically updated teacher network with parameters $\bar\omega$:
\begin{equation}
\begin{split}
b_1&=v_{\bar\omega}(x_s,s,\ell/2;c_t)\\
x'&=x_s+(\ell/2)b_1\\
b_2&=v_{\bar\omega}(x',s+\ell/2,\ell/2;c_t).
\end{split}
\label{eq:shortcut_teacher}
\end{equation}
Here, $\bar\omega$ is a detached, periodically updated teacher snapshot.
The intermediate state $x'$ is obtained through the first half-step, after which the second teacher evaluation estimates the local velocity over the subsequent half-step.
Both evaluations retain the same actuator domain, observation history, and action context, thereby preserving conditioning consistency throughout the teacher rollout.

The two half-step velocities are averaged to obtain a local velocity estimate,
\begin{equation}
\bar b=\tfrac{1}{2}\operatorname{sg}(b_1+b_2).
\label{eq:teacher_velocity}
\end{equation}
Since $F_\omega$ parameterizes the clean endpoint, this velocity estimate is extrapolated over the remaining interval $1-s$ to construct the clean-space shortcut target:
\begin{equation}
y_{\mathrm{sc}}=\operatorname{sg}\!\left[x_s+(1-s)\bar b\right].
\label{eq:shortcut_target}
\end{equation}
Importantly, $\ell$ determines the interval used for the teacher's two-step velocity estimation, whereas $1-s$ defines the extrapolation interval to the clean endpoint.
Thus, the shortcut target is an endpoint estimate based on the teacher's locally averaged velocity, rather than merely the intermediate state after two half-steps.

The generator is optimized using a weighted clean-space regression objective:
\begin{equation}
\mathcal L_{\mathrm{gen}}=\mathbb E\!\left[w(s)\left\|F_\omega(x_s,s,\ell;c_t)-y\right\|_2^2\right].
\label{eq:generator_loss}
\end{equation}
The target is selected according to the requested generation step:
\begin{equation}
y=\begin{cases}
\bar z_{t+1}, & \ell=\ell_{\min}\\
y_{\mathrm{sc}}, & \ell>\ell_{\min}.
\end{cases}
\label{eq:generator_target}
\end{equation}
The time-dependent weighting function is defined as $w(s)=w_0+(1-w_0)s$, where $w_0$ is selected on the validation set with $0<w_0\leq1$.
All shortcut targets are detached from gradient propagation, while gradients are propagated through the student generator.
This objective distills longer-step clean-endpoint predictions from shorter-step teacher evaluations; no formal convergence guarantee is assumed.

During imagination, each predicted transition is generated by integrating $J$ steps from a fresh Gaussian noise sample.
The resulting latent prediction $\hat z_{t+1}$ and its associated action are then appended to the context history for subsequent prediction, so the predicted latent replaces $z_{t+1}$ as the next-latent input at every imagined step.
Here, $K$ denotes the number of imagined transitions, whereas $H$ denotes the length of the observed context window.
These quantities describe distinct aspects of the temporal prediction process.
The generator is used only during training and is omitted at deployment, where the policy operates directly on RGB observations and action history.

\subsection{Physically Grounded Actuator Conditioning}
\label{sec:aci}
\begin{figure}[!t]
\centering
\includegraphics[width=\columnwidth]{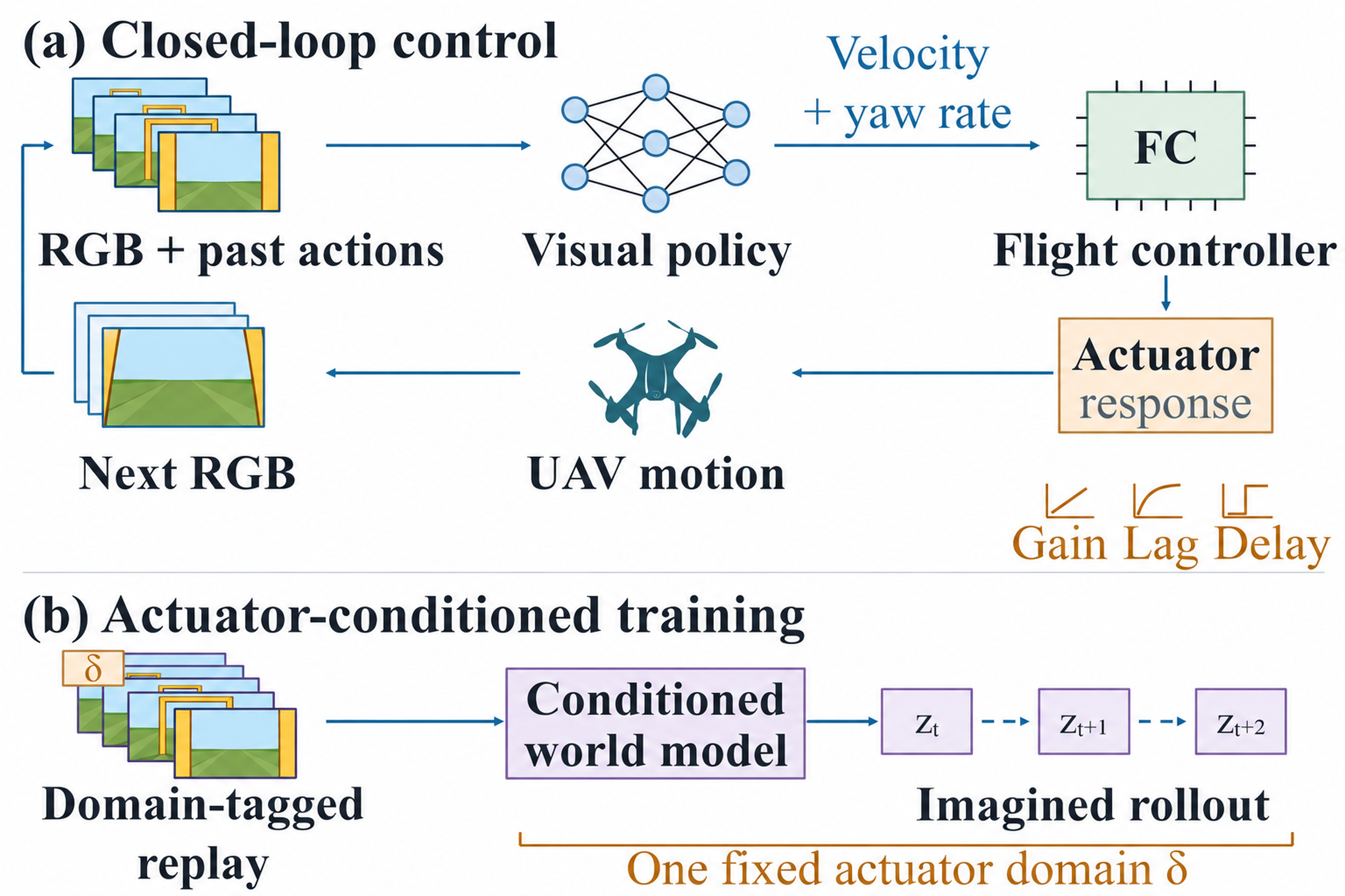}
\caption{\textbf{ACI control and training paths.}
The flight controller maps velocity/yaw-rate commands to actuator inputs.
Domain-tagged replay conditions imagination with one fixed domain per rollout.
Real-flight hardware is described in Section~\ref{sec:hardware}.}
\label{fig:aci_control_chain}
\end{figure}

Figure~\ref{fig:aci_control_chain} illustrates the control and training paths.
ACI varies collective-thrust gain, shared lag, and delay downstream of the velocity-to-thrust controller; lighting, wind, and rotor-specific faults are outside this persistent actuator model.

At actuator tick $n$, define
\begin{equation}
\begin{aligned}
\delta &= (k_T,\tau_T,d)\\
u_n^{\delta} &= k_T\,T^{\mathrm{cmd}}_{n-d}\\
\alpha_\delta &= \exp\!\left(-\Delta t_a/\tau_T\right)\\[6pt]
T_{n+1} &= \operatorname{clip}_{0,T_{\max}}\!\bigl(\alpha_\delta T_n+(1-\alpha_\delta)u_n^{\delta}+\epsilon_n\bigr).
\end{aligned}
\label{eq:actuator_response}
\end{equation}
Positive $k_T,\tau_T$ and integer $d\geq0$ specify gain, lag, and delay in actuator ticks.
The reset delay buffer is initialized consistently, and bounded noise $\epsilon_n$ is sampled at physical ticks.
For velocity commands, Eq.~\ref{eq:actuator_response} acts after velocity-to-thrust conversion; it is an actuator surrogate, not a complete motor model.

Source trajectories retain their known $\delta$ in domain-tagged replay $\mathcal D_\delta$.
Log-scale the positive gain/lag parameters, scale delay by a fixed actuator-tick unit, and normalize using source-only constants $\nu(\delta)$.
A two-layer adapter gives $e_\delta=A_\eta(\nu(\delta))$.
In selected generation blocks it modulates layer-normalized features with bounded scale and shift:
\begin{equation}
\begin{split}
\widetilde f_l={}&
 [1+\beta_l\tanh S_l(e_\delta)]\odot\operatorname{LN}(f_l)\\
 &{}+\beta_l\tanh B_l(e_\delta).
\end{split}
\label{eq:aci_modulation}
\end{equation}
The multiplicative modulation factors remain positive and bounded in $(1-\beta_l,1+\beta_l)$, since $0<\beta_l<1$.
This constraint limits the magnitude of feature modulation rather than directly bounding the physical prediction error.
DRNet and the actor receive no domain code.
Prediction $\hat z_{t+1}=F_{\omega,\eta}(x_s,s,\ell;h_t,a_t,e_\delta)$ uses the same $\delta$ for replay targets, both teacher half-steps, and every imagined transition.
Fresh generation noise varies futures within that domain rather than resampling actuator identity.

Domain-matched replay must cover sampled gains and lags; the actor infers shifts only from RGB/action history, which alone does not establish sim-to-real transfer.

The curriculum samples matching replay/domain pairs from $p_q(\delta)=(1-w_q)p_{\mathrm{nom}}(\delta)+w_qp_{\mathrm{broad}}(\delta)$.
The decreasing schedule uses $w_q=w_{\min}+(w_{\max}-w_{\min})(1-q/Q)_+$, where $q$ counts policy updates, $Q$ is duration, and \mbox{$0<w_{\min}\leq w_{\max}\leq1$}.
Only mixture weights change; broad-domain support is fixed.
Comparisons do not isolate schedule shape from cumulative exposure.

\subsection{Policy Learning in Coherent Imagination}
\label{sec:policy}
Behavior optimization freezes the world model and starts
from a domain-matched replay history. A squashed-Gaussian
actor samples $a_k$, the generator predicts $\hat z_{k+1}$
in the fixed domain, and the heads predict $\hat r_k,\hat c_k$.
Likelihoods include the squashing correction.
For detached target critic $V_{\bar\xi}$, horizon $K\geq1$, and terminal $G_K^\lambda=V_{\bar\xi}(h_K)$, use
\begin{equation}
\begin{split}
G_k^\lambda&=\hat r_k+\gamma\hat c_k\bigl[(1-\lambda)V_{\bar\xi}(h_{k+1})+\lambda G_{k+1}^\lambda\bigr]\\
A_k&=G_k^\lambda-V_\xi(h_k).
\end{split}
\label{eq:lambda_return}
\end{equation}
With $W_0=1$, $W_{k+1}=\gamma\hat c_kW_k$, $\ell_k^\pi=\log\pi_\psi(\operatorname{sg}(a_k)\mid\operatorname{sg}(h_k))$, and policy entropy $\mathcal H_k$, the score-function objective is
\begin{equation}
\mathcal L_\pi=-\mathbb E\sum_{k=0}^{K-1}W_k
 \big[\ell_k^\pi\operatorname{sg}(A_k)+\alpha\mathcal H_k\big].
\label{eq:actor_loss}
\end{equation}
States, sampled actions, returns, and weights are stopped in the score-function update; entropy gradients flow through the policy, and the critic regresses to stopped $G_k^\lambda$.
At $K=0$ the objective is empty; the two zero-horizon baselines instead use PPO~\cite{ppo} and are not horizon-only ablations.

\subsection{Training and Deployment}
\label{sec:training}
Training in Isaac Sim 5.1 has four stages: (1) a PPO teacher~\cite{ppo} collects timestamped episodes with actuator tags, rewards, and terminal labels; (2) the routed codec is trained and frozen; (3) causal dynamics, the adapter, and reward/continuation heads are trained; (4) the world model is frozen and the actor/critic trains from coherent imagined rollouts.
The expert-collected data are not used for actor behavior-cloning pretraining.
Stage (3) minimizes
$\mathcal L_{\mathrm{WM}}=\mathcal L_{\mathrm{gen}}+
\lambda_r\mathcal L_{\mathrm{reward}}+
\lambda_c\mathcal L_{\mathrm{continue}}$,
using squared reward error and binary continuation cross-entropy.
A teacher snapshot is updated only during world-model fitting.

Deployment retains RGB preprocessing, routed encoding, finite-history inference, actor output, and command handoff; the decoder, critic, generator, adapter, and prediction heads are removed.
Complete-path latency quantiles are measured directly across preprocessing, encoding, history inference, actor evaluation, and I/O, including dispatch, transfers, and history recomputation.
Resource accounting distinguishes stored weights from active operators.

\section{Experiments}
\subsection{Experimental Setup}
We evaluate hover, fixed-reference tracking, gate
traversal, and obstacle avoidance using $64\times64$
RGB observations and task-specific checkpoints.
Fixed references define training rewards and
evaluation errors but are excluded from the actor,
which controls the UAV using RGB and command history.
Tracking results do not establish generalization
to arbitrary commanded trajectories.
The comparison includes visual PPO~\cite{ppo},
SAC~\cite{sac}, RSSM Dreamer~\cite{dreamerv3},
and a compact Transformer world model~\cite{dreamer4}.
Observation and action interfaces, simulator conditions,
interaction counts, and test cases matter when
interpreting these comparisons.
Dense and static encoder variants provide
within-architecture routing controls.

Metrics include hover and fixed-reference tracking RMSE,
gate and avoidance success, prediction errors, and
inference latency.
Gate success requires all gates in order without
collision or intervention.
Other gate failures comprise non-timeout failures,
including collisions, missed gates, and safety
interventions.

\subsection{Numerical Comparison}
\label{sec:reported_results}
Table~\ref{tab:main} reports results from five models trained with different random seeds per method, each evaluated on 100 episodes per task; the 100 episodes belong to one independently trained model and are not pooled across seeds.
Each model contributes one task-level aggregate; we report the mean and sample standard deviation of these five aggregates.
The dispersion captures between-model variation together with finite-evaluation variability.
This is a base-architecture comparison and excludes ACI; full ACI transfer is reported separately in Table~\ref{tab:aci_labels}.
Table~\ref{tab:runtime} summarizes online resources.

\begin{table*}[!t]
\centering
\caption{\textbf{Base control without ACI.}
Mean $\pm$ sample SD over five independently trained models, each tested on 100 episodes per task.}
\label{tab:main}
\papertablesetup
\setlength{\tabcolsep}{6pt}
\begin{checkedtable}
\begin{tabularx}{\textwidth}{@{}>{\raggedright\arraybackslash}Xcccccc@{}}
\toprule
Method & \shortstack{Hover RMSE\\(m)$\downarrow$} & \shortstack{Track RMSE\\(m)$\downarrow$} & \shortstack{Gate success\\(\%)$\uparrow$} & \shortstack{Gate failures\\(other, \%)$\downarrow$} & \shortstack{Gate timeout\\(\%)$\downarrow$} & \shortstack{Avoidance\\success (\%)$\uparrow$}\\
\midrule
PPO~\cite{ppo} & $0.085\pm0.004$ & $0.259\pm0.010$ & $87.0\pm2.3$ & $5.2\pm2.3$ & $7.8\pm2.6$ & $82.8\pm2.9$\\
SAC~\cite{sac} & $0.076\pm0.003$ & $0.228\pm0.007$ & $82.4\pm1.7$ & $9.4\pm1.3$ & $8.2\pm2.2$ & $79.8\pm2.9$\\
Dreamer (RSSM)~\cite{dreamerv3} & $0.060\pm0.002$ & $0.180\pm0.006$ & $81.4\pm2.1$ & $11.0\pm1.6$ & $7.6\pm0.5$ & $78.8\pm2.4$\\
Compact Transformer WM~\cite{dreamer4} & $0.055\pm0.002$ & $0.161\pm0.006$ & $84.8\pm1.9$ & $9.2\pm1.9$ & $6.0\pm1.0$ & $81.8\pm2.9$\\
\midrule
\bestrow Tiny Dreamer (w/o ACI) & \best{$0.049\pm0.002$} & \best{$0.139\pm0.005$} & \best{$89.0\pm1.6$} & $7.0\pm1.6$ & \best{$4.0\pm0.7$} & \best{$86.0\pm2.2$}\\
\bottomrule
\end{tabularx}
\end{checkedtable}
\end{table*}

\begin{table*}[!t]
\centering
\caption{\textbf{Online resources.}
Online parameters count weights retained for deployment.
FP16 inference runs on the P450's Jetson Orin NX; latency measures inference, not camera-to-control response time.}
\label{tab:runtime}
\papertablesetup
\setlength{\tabcolsep}{6pt}
\begin{checkedtable}
\begin{tabular*}{\textwidth}{@{\extracolsep{\fill}}lcccccc@{}}
\toprule
Method & \shortstack{Online params\\(M)} & \shortstack{Active compute\\(GFLOPs)} & \shortstack{Median\\(ms)} & \shortstack{P95\\(ms)} & \shortstack{P99\\(ms)} & \shortstack{Peak memory\\(MB)}\\
\midrule
PPO~\cite{ppo} & 0.391 & 0.24 & 7.6 & 10.2 & 12.4 & 410\\
SAC~\cite{sac} & 0.523 & 0.31 & 9.1 & 12.0 & 14.7 & 500\\
Dreamer (RSSM)~\cite{dreamerv3} & 1.511 & 1.65 & 33.6 & 44.0 & 53.8 & 1280\\
Compact Transformer WM~\cite{dreamer4} & 0.718 & 0.94 & 23.2 & 31.5 & 38.6 & 860\\
\midrule
Tiny Dreamer (all branches) & 0.520 & 0.53 & 15.1 & 20.2 & 24.5 & 540\\
Tiny Dreamer (routed) & 0.520 & 0.36 & 12.9 & 17.4 & 21.2 & 520\\
\bottomrule
\end{tabular*}
\end{checkedtable}
\end{table*}

Tiny Dreamer improves gate success by 4.2 percentage points and tracking RMSE by 0.022\,m over the compact Transformer~\cite{dreamer4} in the reported base-architecture comparison.
Routing reduces active compute by 32.1\% and P95 inference latency by 13.9\% relative to Tiny Dreamer with all branches active.

Routed Tiny Dreamer uses less compute and runs faster than the evaluated model-based baselines, while PPO~\cite{ppo} and SAC~\cite{sac} retain lower deployment costs (Table~\ref{tab:runtime}). Routing, history processing, and dispatch can add latency beyond what active GFLOPs predict. FP16 inference measurements on the Jetson Orin NX exclude the generator and critic.

\subsection{Ablation Study}
\begin{figure}[tbp]
\centering
\includegraphics[width=\columnwidth]{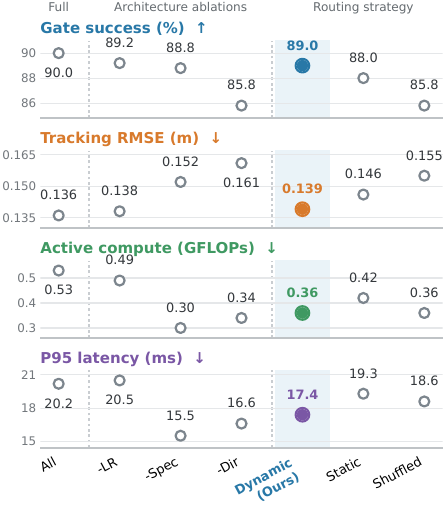}
\caption{\textbf{DRNet ablation and routing comparison.}
Each row uses an independent vertical scale and reports raw values.
Higher gate success is better, whereas lower tracking RMSE, active compute, and P95 latency are better. The shaded column highlights dynamic routing (ours).
}
\label{fig:drnet_ablation}
\end{figure}

Figure~\ref{fig:drnet_ablation} compares encoder branches and routing, while Table~\ref{tab:temporal_ablation} compares temporal settings.
The reference uses $H=8$ and $K=15$, matching Tables~\ref{tab:main} and~\ref{tab:runtime}.

\textbf{\mbox{Static Routing Control.}} The static encoder executes a fixed subset of $B_{\mathrm{act}}$ branches for every input and uses input-independent fusion weights.
The dynamic variant selects the same number of branches based on the input.
Figure~\ref{fig:drnet_ablation} shows 0.36 GFLOPs for dynamic routing versus 0.42 GFLOPs for static routing; their P95 latencies are 17.4 and 19.3\,ms, respectively.
\textbf{\mbox{Input-Dependent Routing.}} We compare dynamic routing with shuffled routing, which reassigns routing decisions across observations while keeping network parameters unchanged. At equal active computation (0.36 GFLOPs), dynamic routing improves gate success from 85.8\% to 89.0\% and reduces tracking RMSE from 0.155\,m to 0.139\,m. These results support the benefit of input-dependent branch selection.

\textbf{\mbox{Lightweight-Encoder Comparison.}} Only the visual backbone changes; a projection restores the same latent width, and the world model and actor--critic are retrained from scratch with identical data, update counts, and interaction budgets.
All variants use $64\times64$ RGB, five paired seeds, and 100 matched episodes per task; FP16 full-path resources are profiled on the same Jetson Orin NX.
Table~\ref{tab:encoder_backbone} compares the five visual encoders under this protocol.

\begin{table*}[!t]
\centering
\caption{\textbf{Lightweight-encoder comparison.}
Online resource columns follow Table~\ref{tab:runtime}; gate success and tracking RMSE are five-seed mean $\pm$ sample SD. Named backbone widths are retained, so parameter budgets differ.}
\label{tab:encoder_backbone}
\papertablesetup
\setlength{\tabcolsep}{6pt}
\begin{checkedtable}
\begin{tabular*}{\textwidth}{@{\extracolsep{\fill}}lccccccc@{}}
\toprule
Visual encoder & \shortstack{Online\\params (M)} & \shortstack{Active\\GFLOPs} & \shortstack{P95\\(ms)} & \shortstack{P99\\(ms)} & \shortstack{Peak\\MB} & \shortstack{Gate\\(\%)$\uparrow$} & \shortstack{Track RMSE\\(m)$\downarrow$}\\
\midrule
MobileNetV3-Small\cite{mobilenetv3} & 2.900 & 0.44 & 18.2 & 22.4 & 615 & $87.0\pm1.9$ & $0.149\pm0.006$\\
ShuffleNetV2 0.5$\times$~\cite{shufflenetv2} & 1.750 & 0.29 & 16.1 & 20.0 & 555 & $85.8\pm2.1$ & $0.156\pm0.007$\\
DRNet Static & 0.520 & 0.42 & 19.3 & 23.4 & 525 & $88.0\pm1.8$ & $0.146\pm0.006$\\
DRNet All-Branch & 0.520 & 0.53 & 20.2 & 24.5 & 540 & $90.0\pm1.7$ & $0.136\pm0.005$\\
\bestrow DRNet Dynamic & \best{0.520} & \best{0.36} & \best{17.4} & \best{21.2} & \best{520} & \best{$89.0\pm1.6$} & \best{$0.139\pm0.005$}\\
\bottomrule
\end{tabular*}
\end{checkedtable}
\end{table*}

\begin{table}[!t]
\centering
\caption{\textbf{Temporal comparison.}
$H$ is observed context; $K$ is imagination horizon; RMSE is tracking error.
Zero-horizon variants are alternative baselines rather than horizon-only ablations.}
\label{tab:temporal_ablation}
\papertablesetup
\setlength{\tabcolsep}{4pt}
\begin{checkedtable}
\begin{tabularx}{\columnwidth}{@{}>{\raggedright\arraybackslash}Xccccc@{}}
\toprule
Variant & $H$ & $K$ & \shortstack{Gate\\(\%)$\uparrow$} & \shortstack{RMSE\\(m)$\downarrow$} & \shortstack{P95\\(ms)$\downarrow$}\\
\midrule
Single-frame actor & 1 & 0 & 67.8 & 0.237 & 9.6\\
History, no imagination & 8 & 0 & 79.3 & 0.182 & 13.1\\
Short context & 4 & 15 & 85.4 & 0.158 & 14.9\\
\bestrow Reference & 8 & 15 & \best{89.0} & \best{0.139} & \best{17.4}\\
Long context & 16 & 15 & 90.2 & 0.135 & 24.8\\
\bottomrule
\end{tabularx}
\end{checkedtable}
\end{table}

At $K=15$, increasing $H$ from 4 to 8 raises gate success from 85.4\% to 89.0\% and P95 latency from 14.9 to 17.4\,ms; $H=16$ reaches 90.2\% at 24.8\,ms.
Both $K=0$ actors use PPO~\cite{ppo}, so comparison with imagined-policy learning changes the training objective as well as horizon.

\subsection{Real-Platform Setup and Transfer Protocol}
\label{sec:hardware}
\begin{figure}[!t]
\centering
\includegraphics[width=\columnwidth,trim=0bp 1164bp 0bp 600bp,clip]{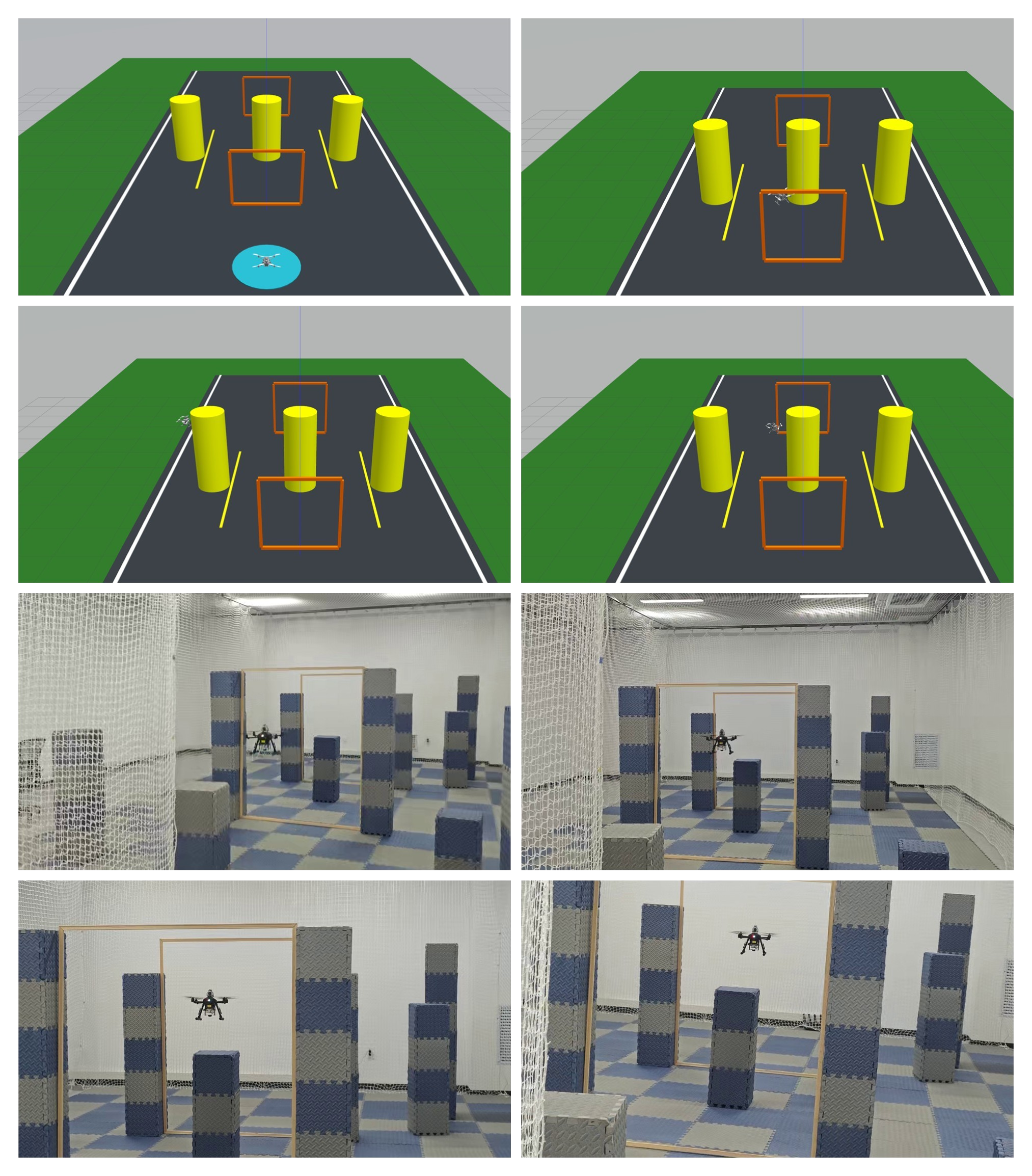}
\par\vspace{1.5pt}
\includegraphics[width=\columnwidth,trim=0bp 36bp 0bp 1728bp,clip]{images/gazebo_real_flight_sequences_8panel_single_column.png}
\caption{\textbf{Qualitative flight sequences.}
The first row shows two selected Gazebo flight stages; the second row shows two real-flight gate-traversal stages.}
\label{fig:flight_sequences}
\end{figure}
Real-flight validation uses a P450 quadrotor with a 16-GB Jetson Orin NX (Ubuntu 20.04.6; JetPack 5.1.4/L4T 35.6.0), a RealSense D435i RGB camera (640$\times$480; 30~Hz configured and 29.96~Hz logged), and a Livox MID360 lidar.
FAST-LIO provides state feedback to the low-level controller and trajectories for evaluation, while its estimates do not enter the RGB/history actor.
Through ROS Noetic, Prometheus, and MAVROS, the frozen actor sends body-frame velocity and yaw-rate commands at 20~Hz to PX4, which stabilizes the UAV.
Figure~\ref{fig:flight_sequences} presents qualitative flight sequences in Gazebo and on the real platform.
\textbf{\mbox{Perturbation Protocol.}} Upstream of PX4, we perturb P450 velocity and yaw-rate commands with gain scaling ($0.8\times$ or $1.2\times$), first-order lag (0.10 or 0.20\,s), or a 100\,ms delay. B1 and B7 each fly 10 trials per setting with matched references and initial conditions. Settings are fixed per flight. We log applied commands and perturbations and retain PX4 stabilization and safety intervention. These command-path perturbations test closed-loop robustness but do not directly reproduce physical motor or thrust variations.

\subsection{Actuator-Shift Evaluation}
\label{sec:actuator_evaluation}

\textbf{\mbox{Prediction Evaluation.}} Data are partitioned by complete trajectories; multi-step predictions recursively append predicted latents to the history.
The proposed source ranges are $k_T\in[0.9,1.1]$, $\tau_T/\tau_0\in[0.75,1.25]$, and $d\in\{0,1\}$; proposed stresses use gains 0.8/1.2, lag ratios 0.5/1.5, and delay 2.
These simulator settings are separate from real-flight target errors.
For prediction horizon $k$, endpoint latent MSE is defined as
\begin{equation}
\operatorname{MSE}_k=\frac{1}{ND}\sum_{i=1}^{N}
\left\|\hat z_{t+k}^{(i)}-\operatorname{sg}(f_\theta(o_{t+k}^{(i)}))\right\|_2^2.
\label{eq:multistep_mse}
\end{equation}
Here $N$ is the sequence count and $D$ is the latent dimension.
This metric measures error at horizon $k$, rather than averaging over horizons.
Reward MAE measures absolute reward-prediction error; the continuation Brier score measures squared probability error against the binary continuation target.
At 20 steps, B7 reduces latent MSE to 0.169 from B1's 0.294 (B6: 0.181). B6 has the lowest one-step MSE (0.018 versus 0.019 for B7), while B7 has the lowest reward MAE (0.070) and continuation Brier score (0.052), versus 0.094 and 0.071 for B1.

\textbf{\mbox{Control Evaluation.}} B0 uses nominal replay; B1 uses actuator-randomized replay without conditioning.
B2/B3 add action/latent noise; B4 uses rollout-fixed dropout.
B5 directly conditions on $\delta$; B6 uses a fixed-mixture adapter; B7 adds curriculum scheduling.
B5--B7 retain one domain per rollout.
B1--B7 share data, interfaces, budgets, and test seeds; hyperparameters use validation shifts.

\textbf{\mbox{Transfer Evaluation.}} Frozen Isaac Sim models are evaluated zero-shot in Gazebo and on the P450 (Section~\ref{sec:hardware}).
Table~\ref{tab:aci_labels} reports tracking RMSE and signed gap $G_m=E_m^{\mathrm{P450}}-E_m^{\mathrm{Gazebo}}$; the absolute gap reduction relative to nominal is $\Delta G_m=G_{\mathrm{B0}}-G_m$ (m).
Differences require matched tasks, frames, and evaluation windows; absolute flight errors are also reported.

\begin{table*}[!tbp]
\centering
\caption{\textbf{Effect of actuator conditioning on zero-shot transfer.}
Gazebo and Flight report tracking RMSE; Gap is their difference (m).
The 95\% CI is a Student's $t$ interval over the $n$ flights of each condition.}
\label{tab:aci_labels}
\papertablesetup
\setlength{\tabcolsep}{6pt}
\begin{checkedtable}
\begin{tabular*}{\textwidth}{@{\extracolsep{\fill}}lccccccc@{}}
\toprule
Condition & \shortstack{Gazebo\\(m)} & \shortstack{Flight\\(m)} & \shortstack{Gap\\(m)} & $n$ & \shortstack{SD\\(m)} & \shortstack{Variance\\(m$^2$)} & \shortstack{95\% CI\\(m)} \\
\midrule
B0 Nominal & 0.139 & 0.264 & 0.125 & 10 & 0.024 & 0.000576 & [0.247, 0.281] \\
\midrule
B1 Unconditioned & 0.153 & 0.218 & 0.065 & 10 & 0.021 & 0.000441 & [0.203, 0.233] \\
B2 Action noise & 0.157 & 0.229 & 0.072 & 10 & 0.023 & 0.000529 & [0.213, 0.245] \\
B3 Latent noise & 0.159 & 0.236 & 0.077 & 10 & 0.025 & 0.000625 & [0.218, 0.254] \\
B4 Model dropout & 0.156 & 0.224 & 0.068 & 10 & 0.022 & 0.000484 & [0.208, 0.240] \\
\midrule
B5 Direct code & 0.151 & 0.205 & 0.054 & 10 & 0.020 & 0.000400 & [0.191, 0.219] \\
B6 Correct labels & 0.150 & 0.190 & 0.040 & 10 & 0.018 & 0.000324 & [0.177, 0.203] \\
Shuffled labels & 0.155 & 0.221 & 0.066 & 10 & 0.022 & 0.000484 & [0.205, 0.237] \\
Random codes & 0.156 & 0.224 & 0.068 & 10 & 0.023 & 0.000529 & [0.208, 0.240] \\
\bestrow B7 Scheduled ACI & 0.146 & \best{0.181} & \best{0.035} & 10 & \best{0.017} & \best{0.000289} & \best{[0.169, 0.193]} \\
\bottomrule
\end{tabular*}
\end{checkedtable}
\end{table*}

\begin{figure}[!htbp]
\centering
\includegraphics[width=\columnwidth]{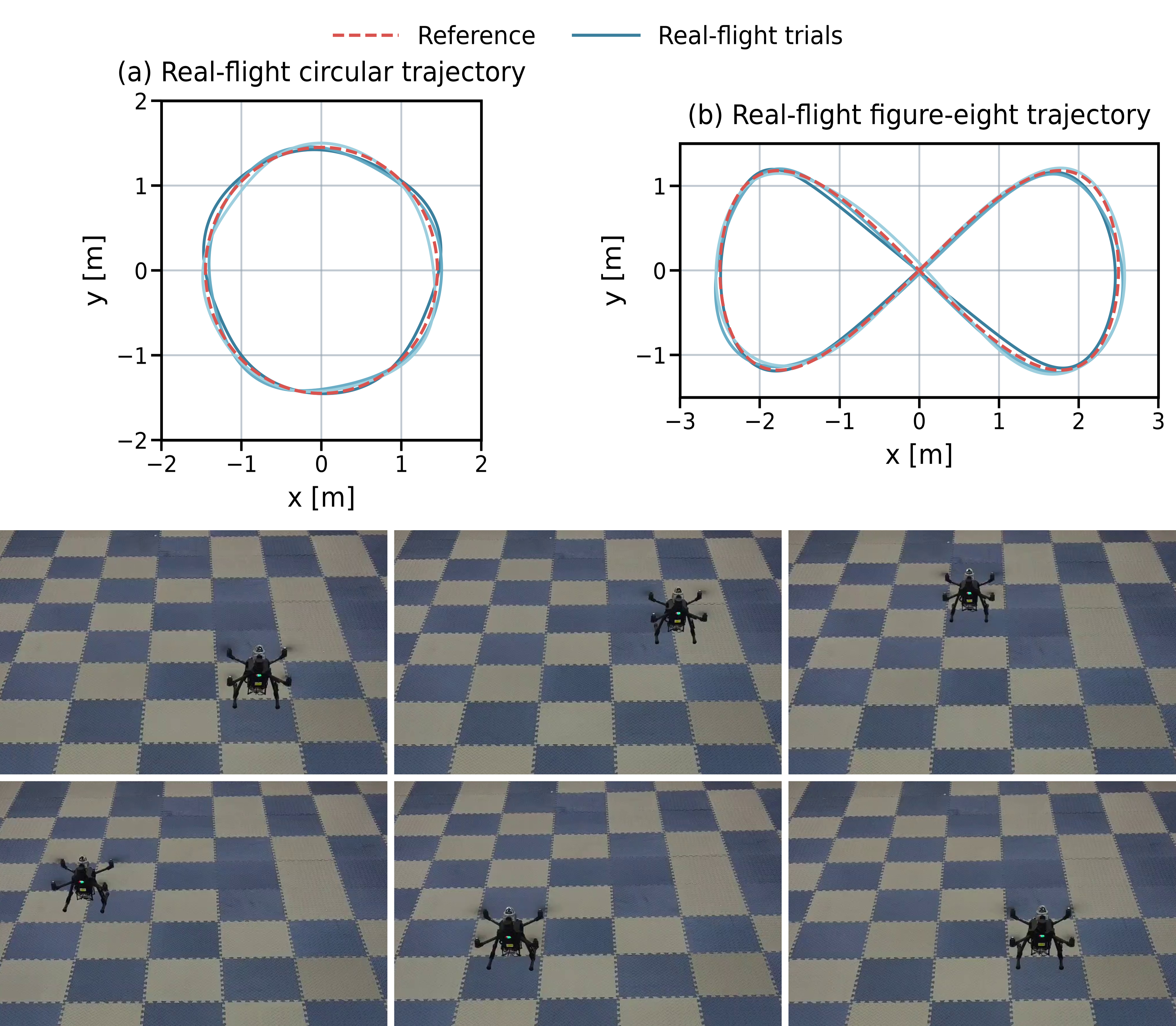}
\caption{\textbf{Real-flight trajectories and tracking frames.}
The top panels compare reference (red dashed) and
real-flight (blue solid) trajectories for circular
and figure-eight tracking. The bottom panels show
representative frames from UAV experiments.}

\label{fig:real_flight_tracking}
\end{figure}

Table~\ref{tab:aci_labels} reports zero-shot transfer for all variants on the nominal P450 platform.
Scheduled ACI reaches 0.181\,m flight RMSE and outperforms unconditioned randomization (0.218\,m).
Correct labels yield 0.190\,m RMSE, versus 0.221\,m and 0.224\,m for shuffled labels and random codes, supporting physically informative conditioning.
For 10 real flights per condition, we report sample SD, variance, and 95\% Student's $t$ intervals for the mean. Figure~\ref{fig:real_flight_tracking} illustrates real-flight tracking.
Under the tested P450 command-path perturbations, scheduled ACI reduces mean tracking RMSE by 17.3--20.5\% versus unconditioned randomization (Table~\ref{tab:p450_command_shifts}).
Each Gain or Lag row in that table aggregates the two settings of one perturbation for a single policy, so its 20 flights equal $2\times10$ flights per setting and match the protocol above, whereas the nominal and delay rows use 10 flights each.

\begin{table}[!t]
\centering
\caption{\textbf{P450 command-path perturbations.} Per-flight RMSE (m, mean $\pm$ sample SD). Per policy, Gain and Lag pool 20 flights each (10/setting), Nominal and Delay 10 each. Reduction vs. B1.}
\label{tab:p450_command_shifts}
\papertablesetup
\setlength{\tabcolsep}{6pt}
\renewcommand{\arraystretch}{1.0}
\begin{tabularx}{\columnwidth}{@{}Xccc@{}}
\toprule
Condition & B1 $\downarrow$ & B7 $\downarrow$ & Reduction $\uparrow$ \\
\midrule
Nominal & $0.218\pm0.021$ & $0.181\pm0.017$ & 17.0\% \\
Gain shift & $0.249\pm0.026$ & $0.198\pm0.020$ & 20.5\% \\
Lag shift & $0.263\pm0.028$ & $0.211\pm0.023$ & 19.8\% \\
Delay shift & $0.284\pm0.030$ & $0.235\pm0.026$ & 17.3\% \\
\bottomrule
\end{tabularx}
\end{table}

\subsection{Actuator-Conditioning Diagnostic}
We isolate label information and temporal consistency in scalar thrust-response simulations of Eq.~\ref{eq:actuator_response}, with normalized thrust clipped to $[0,1]$, $\Delta t_a=0.02$\,s, and no process noise.
Independent training and test sets each contain 100 trajectories of 64 ticks; commands in $[0.1,0.9]$ are held for four ticks, while $k_T\sim U[0.8,1.2]$, $\tau_T\sim U[0.05,0.15]$\,s, and $d\sim U\{0,1,2\}$ remain fixed per trajectory.
Across five seeds, all variants use the same 289-parameter tanh predictor trained for 1,200 Adam updates, and compare correct conditions with zero, shuffled, random-code, masked, and per-tick-resampled conditions.
Correct labels reduce 20-step thrust RMSE from 0.0790 to 0.0302 relative to zero conditions; shuffled labels and per-tick resampling increase error, while delay masking improves one-step accuracy but worsens 20-step prediction.

\subsection{Component and Policy-Training Comparison}
Table~\ref{tab:component_ablations} compares the ACI adapter, curriculum, and visual-policy training under a test configuration different from those of Tables~\ref{tab:main} and~\ref{tab:aci_labels}.
We evaluate fixed-reference tracking and gate traversal in a separate Isaac Sim actuator-shift suite. Each episode fixes one of the four settings of $k_T\in\{0.8,1.2\}$ and $\tau_T/\tau_0\in\{0.5,1.5\}$ at $d=2$; each variant sees 100 episodes per task (25 per setting) with matched initial states. Reported values are five-seed means with sample SDs across seed-level aggregates. All three blocks share one full-configuration result (92.0\% gate success), and the visual-policy comparison also changes the training objective.

\begin{table}[!htbp]
\centering
\caption{\textbf{Ablations of the ACI adapter, curriculum, and visual-policy training.}
$^\dagger$Shared reference across all three blocks.}
\label{tab:component_ablations}
\papertablesetup
\setlength{\tabcolsep}{6pt}
\begin{checkedtable}
\begin{tabularx}{\columnwidth}{@{}>{\raggedright\arraybackslash}Xcc@{}}
\toprule
Variant & \shortstack{Track\\RMSE (m)$\downarrow$} & \shortstack{Gate\\success (\%)$\uparrow$} \\
\midrule
\multicolumn{3}{@{}l}{\emph{(a) ACI adapter}}\\
Unconditioned WM & $0.270 \pm 0.009$ & $86.0 \pm 2.7$ \\
Direct domain code & $0.240 \pm 0.007$ & $89.0 \pm 2.2$ \\
\bestrow Full ACI adapter$^\dagger$ & \best{$0.220 \pm 0.005$} & \best{$92.0 \pm 1.6$} \\
\midrule
\multicolumn{3}{@{}l}{\emph{(b) Curriculum}}\\
Fixed mixture & $0.240 \pm 0.008$ & $89.0 \pm 2.2$ \\
Warm-up schedule & $0.230 \pm 0.007$ & $90.0 \pm 2.2$ \\
\bestrow Decreasing schedule$^\dagger$ & \best{$0.220 \pm 0.005$} & \best{$92.0 \pm 1.6$} \\
\midrule
\multicolumn{3}{@{}l}{\emph{(c) Visual policy}}\\
Single-frame PPO~\cite{ppo} & $0.310 \pm 0.015$ & $80.0 \pm 3.2$ \\
History PPO~\cite{ppo} & $0.280 \pm 0.010$ & $85.0 \pm 2.1$ \\
\bestrow Full imagined-policy training$^\dagger$ & \best{$0.220 \pm 0.005$} & \best{$92.0 \pm 1.6$} \\
\bottomrule
\end{tabularx}
\end{checkedtable}
\end{table}

\section{Conclusion and Outlook}
\label{sec:conclusion}
Tiny Dreamer combines routed RGB encoding,
finite-context dynamics, and actuator-conditioned
imagination for UAV control. DRNet reduces active
compute by 32.1\% and P95 latency by 13.9\%
compared with all-branch execution. The deployed
policy uses RGB and command history without online
generation. Scheduled ACI lowers zero-shot P450 tracking error relative to unconditioned randomization under nominal and tested command-path conditions.

\textbf{Limitations and Future Work.}
Robustness to physical actuator changes remains
untested. ACI covers persistent gain, lag, and delay,
while broader dynamics and long-horizon prediction
remain future challenges.

\section*{LLM Usage Statement}
We used OpenAI ChatGPT to assist with English editing, grammatical polishing, and improving the clarity and readability of the manuscript. All AI-assisted suggestions were carefully reviewed and revised by the authors to ensure scientific accuracy and consistency with the intended meaning. The authors retain full responsibility for the research methodology, technical content, experimental results and their interpretation, figures, tables, references, and the final submitted manuscript.

{\footnotesize
\setlength{\bibsep}{0pt plus 0.5pt}
\bibliographystyle{ieeenat_fullname}
\bibliography{main1}
}

\end{document}